\documentclass{article}

\usepackage{arxiv}

\usepackage[utf8]{inputenc} 
\usepackage[T1]{fontenc}    
\usepackage{hyperref}       
\usepackage{url}            
\usepackage{booktabs}       
\usepackage{amsfonts}       
\usepackage{nicefrac}       
\usepackage{microtype}      
\usepackage{lipsum}
\usepackage{siunitx}
\usepackage[lofdepth,lotdepth]{subfig}
\usepackage{amsmath}
\usepackage{graphicx}
\graphicspath{{./images/}}
\title{Autonomous Braking and Throttle System: A Deep Reinforcement learning approach for naturalistic driving}

\author{
  Varshit S.~Dubey \\
  Dept. of Electronics and Telecommunication Engg.\\
  College of Engineering Pune\\
  Pune, India \\
  \texttt{dubeyvs16.extc@coep.ac.in} \\
  \And
 Karan .~Agrawal \\
  Dept. of Electronics and Telecommunication Engg.\\
  College of Engineering Pune\\
  Pune, India \\
  \texttt{agrawalkr16.extc@coep.ac.in} \\
  \And
  Ruhshad .~Kasad \\
  Dept. of Electronics and Telecommunication Engg.\\
  College of Engineering Pune\\
  Pune, India \\
  \texttt{kasadrr16.extc@coep.ac.in} \\
}

\begin{document}
\maketitle

\begin{abstract}
Autonomous Braking and Throttle control is key in developing safe driving systems for the future. There exists a need for autonomous vehicles to negotiate a multi-agent environment while ensuring safety and comfort. 
A Deep Reinforcement Learning based autonomous throttle and braking system is presented. For each time step, the proposed system makes a decision to apply the brake or throttle. The throttle and brake are modelled as continuous action space values. We demonstrate 2 scenarios where there is a need for a sophisticated braking and throttle system, i.e when there is a static obstacle in front of our agent like a car, stop sign. The second scenario consists of 2 vehicles approaching an intersection. The policies for brake and throttle control are learned through computer simulation using Deep deterministic policy gradients. The experiment shows that the system not only avoids a collision, but also it ensures that there is smooth change in the values of throttle/brake as it gets out of the emergency situation and abides by the speed regulations, i.e the system resembles human driving.
\end{abstract}

\keywords{Autonomous systems \and Collision Avoidance system \and Deep Reinforcement learning \and Naturalistic driving \and Simulation}

\section{Introduction}
The past decade has seen exponential growth in technologies associated with autonomous vehicles. Major technology companies and vehicle manufacturers have invested in autonomous vehicles. Autonomous vehicles are exposed to countless dynamic scenarios involving multiple agents such as vehicles, pedestrians and signs. Autonomous vehicles require reliable control systems to deal with such uncertainty. Active safety measures such as collision avoidance systems are becoming increasingly popular on cars. 

Autonomous Braking systems are already available on most cars. Such systems detect an impending crash and take corrective action in time to avoid accidents. These systems work independent of the driver and apply brakes to decelerate and stop the vehicle avoiding incidents or reduce impact. Autonomous Braking systems work in the narrow zone between braking early to avoid accidents and braking late to conserve performance.

Advanced cruise control systems use a combination of throttle and braking to maintain a constant velocity following a vehicle. Such features elevate vehicles to Level 1 of driving automation. However, they fall short of creating a high level of automation.

Traditionally rule-based control systems are unable to adapt to the dynamic environment autonomous vehicles have to operate in. Rule-based systems specify a distance threshold or a speed threshold and often require human assistance in complex situations. An intelligent system is required to cover a wide spectrum of scenarios.

Intelligent algorithms are required to make such systems viable. Autonomous Throttle and Braking Systems need to be accurate in a high percentage of scenarios. False-alarm scenarios erode driver confidence and can be dangerous. Such systems cannot be programmed for the near infinite scenarios which may occur, therefore there exists a need for such systems to be able to learn and adapt. Driving and specifically, scenarios requiring emergency brake action can be considered a multi-agent problem with only a partial observable space at any given instant. 

The death of a pedestrian due to collision with an Uber self-driving car highlights the need for evolved braking and throttle systems. Despite detecting the pedestrian early, the system failed to act and exceeded its assured clear distance ahead.

Reinforcement Learning is a methodology allowing an agent to learn by interacting and participating in an environment. The aim of using such methods is to obtain an optimal policy which maximizes the reward when the agent interacts with the environment. Policy Gradient methods optimizes the policy directly. The ability of Policy Gradient methods to learn stochastic policy is integral in implementing a throttle and braking control system. \cite{10.5555/3312046}

In this paper a throttle and braking control system is developed using Deep Reinforcement Learning. Specifically, policy gradient methods are utilized to dictate discrete brake and throttle inputs. The performance of the system is tested in two scenarios which are simulated in CARLA simulator. \cite{Dosovitskiy17}

The rest of the paper is organized as follows. Section 2, will include a brief review of research into braking control systems implemented using Deep Reinforcement Learning.Section 3 presents the scenarios and simulation environment to test the proposed braking and throttle system. Section 4 contains details regarding the reinforcement learning methods used and the proposed reward function. Section 5 contains the result of conducted experiments and Section 6 discusses future work.
\section{Related Work}

Simulation is integral in the development of autonomous vehicles. Developing algorithms to navigate in dense urban environments has been a key focus area. Dosovitskiy \cite{Dosovitskiy17} noted that there are obstacles to real-life research such as government regulation, logistical difficulties and the possibility of accidents. Training an agent for the possibility of a rare event such as jaywalkers is also a  necessity. One way to resolve the testing and training of autonomous vehicles is the use of simulation software. CARLA open-source simulator provides an urban environment for simulation with the ability to generate pedestrians, vehicles and common road signs. The ability to gather data such as velocity, coordinates and distance to other vehicles reduces the effort in acquiring specialized data. Moreover, the ability to set up an environment for scenarios to train is beneficial.

According to 2007 statistics for road accidents in Germany, 14 \% of accidents occur between a vehicle following a moving vehicle ahead or a static vehicle. Analysis of frontal collisions with stationary obstacles exhibited that 21\% of the drivers do not brake at all prior to the accident. Kaempchen \cite{5161310} proposes an emergency braking system which generates potentially dangerous scenarios after detecting an object or vehicle. The paper analyses all possible trajectories of the host and object, while calculating the possibility of a collision. Specifically, it provides insight in two common scenarios, rear-end collision and collisions at cross-sections.

Traditionally rule-based systems were utilized to create braking and throttle control systems. Such systems had limited scope and needed specific protocols for specific scenarios and would be handicapped in the real-world. Chae \cite{8317839} concludes that a learning-based system is likely to achieve better results in adapting to new and dynamic scenarios than rule-based systems. The paper realizes braking control systems as a discrete Markov Decision Process. Moreover the braking decision is initiated at a trigger point which is calculated. 

However such an approach is counterintuitive to real-life driving since, braking decision can be decided at any point during the approach. The author has used DQN to design a braking system with 4 discrete action values, $a_{0}$, $a_{low}$, $a_{mid}$, $a_{high}$ \cite{8317839}. The main drawback of this approach is that DQN works on discrete state space whereas braking is continuous value in the real world, which makes it difficult for DQN to find optimal policy.  This paper considers the continuous space of the brake/throttle and uses DDPG which is similar to DQN, but uses different subroutines to find the optimal policy and works for continuous action spaces. The details of DDPG will be explained in the subsequent section. 

Reinforcement Learning considers the problem of an agent trying to achieve a goal while interacting with an unknown environment, therefore it is well-suited to deal with the problem of a vehicle navigating in traffic. Talpeart \cite{visapp19} focuses on teaching an agent to establish a policy while interacting with an environment to maximize a numerical reward.

Vasquez \cite{8916912}. focuses on emergency braking to avoid accidents with pedestrians. A key aspect of the paper is modelling comfort as reduction of jerk while braking and creating a naturalistic dataset which closely mimics pedestrian behaviour. Early-braking methods combined with the throttle action as discussed in this paper are designed to reduce jerk. However the idea of creating datasets to mimic real-world behaviour can be extended to driving policies in order to make policies more naturalistic.

Y.Fu \cite{9067008}. examines Deep Deterministic Policy Gradient (DDPG) algorithm based on Actor-Critic infrastructure  to reduce the difficulty of learning in more complex environments. The paper proposes that a reward function should accommodate multiple objectives, such as safety, comfort. A DDPG system consisting of only braking action outperforms similar braking systems based on fuzzy logic and DQN. Another key criteria examined is reducing the probability of exceeding comfortable deceleration or experiencing jerk. On the same count, DDPG systems exceed comfortable deceleration limits on fewer occasions than DQN or fuzzy logic systems.

There has been an increased interest in using a DRL method for emergency braking scenarios. The reward function penalizes early braking and collisions with other vehicles and pedestrians. A Deep Deterministic Policy Gradient method allows the agent to rapidly discover optimal policies in simulation by using stochastic policies for exploration to learn a deterministic policy.\cite{8569222}

Previous research largely focuses on braking action as a final resort to either prevent or mitigate collisions. Such measures are limited in scope. Braking action along with throttle control is more closely related with real-world driving practices. This paper focuses on 2 scenarios. The first scenario tests the control systems braking ability while scenario 2, works with throttle action necessary after an emergency situation.

\section{System Description}
\begin{figure}[h!]
    \centering
    \includegraphics[scale=0.5]{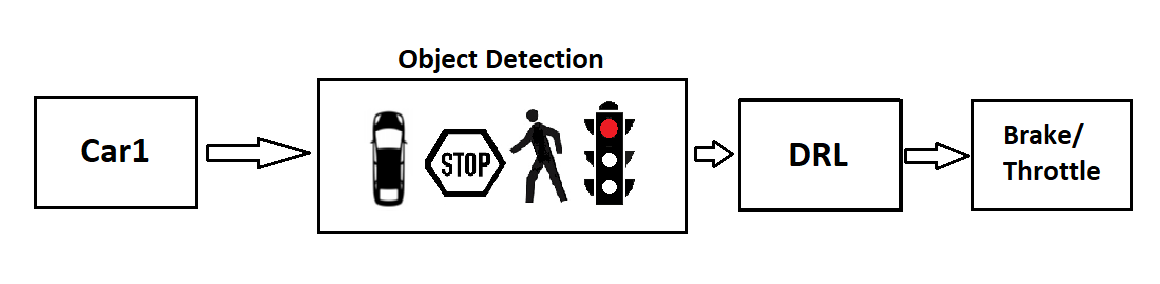}
    \caption{High level overview of Autonomous Braking/throttle system}
    \label{fig:figure}
\end{figure}

Society of Automotive Engineering defines 6 levels of driving automation from level 0 to level 5. Level 5 of driving automation can drive a vehicle anywhere under all possible conditions. A driver is not required to take control at any point and vehicles provide a safe and comfortable journey.
The biggest challenge in achieving driverless cars is safety. A broad overview of braking and throttle system using reinforcement learning is shown in 
figure \ref{fig:figure}. A car should detect the surrounding objects and take decisions such as deciding velocity,steering angle and braking to avoid accidents. An object in the surrounding environment of the vehicle could be a static obstacle, pedestrian, a stop sign or a moving car.
Once the object is detected, different parameters of the car and the object will be passed into the neural network \cite{10.5555/3086952} and then via reinforcement learning an output will be generated. The output can be a throttle or brake value which should be applied by the car as shown in figure \ref{fig:figure}. The parameters to be passed into the network as input can be initial velocity of car, relative position of the object to the car, relative velocity etc. 
\newline
In this paper we are considering 2 common scenarios, where there is a need for effective braking and throttle system to avoid accidents.

\begin{figure}[h!]
    \centering
    \includegraphics[scale=0.7]{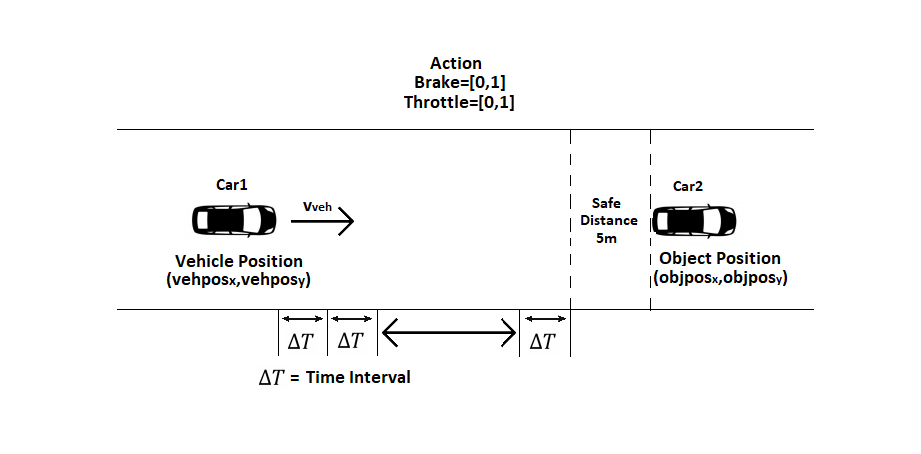}
    \caption{System description for Scenario 1}
    \label{fig:figure1}
\end{figure}

\begin{figure}[h]
    \centering
    \includegraphics[scale=0.6]{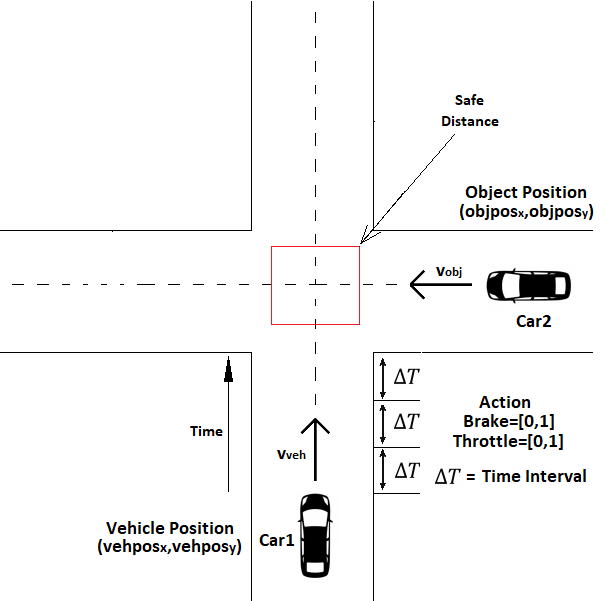}
    \caption{System description for Scenario 2}
    \label{fig:figure2}
\end{figure}

\subsection{Scenario 1}
The first scenario is shown in figure \ref{fig:figure1}. There is a static obstacle in front of the car at some distance. The static obstacle can be another vehicle, stop sign etc. The autonomous agent is moving with the velocity of $v_{veh}$ from a position $(vehpos_x, vehpos_y)$. As soon as the static object is detected, the vehicle receives the relative position of the object, i.e 
$(objpos_x-vehpos_x, objpos_y-vehpos_y)$. From the coordinates, the relative distance and direction from the agent vehicle is calculated. Utilizing this information, the agent decides whether to apply brake or throttle at each time step. The brake and throttle values are between $[0,1]$.

\subsection{Scenario 2}
In the second scenario, 2 cars are approaching the intersection. We consider that the other car approaching the intersection is not abiding by traffic rules. This creates a scenario in which an agent is required to make a quick decision to apply brakes to avoid an accident. After we avoid this emergency situation, we can again apply throttle to move forward. We will try to make our autonomous agent learn this policy. Here, the autonomous agent and the object are moving with velocities $v_{veh}$ and $v_{obj}$ respectively. When the episode starts, the agent starts from position $(vehpos_x, vehpos_y)$ and the other car starts from $(objpos_x, objpos_y)$. We assume that at a certain time, using sensors, our agent gets information about the object approaching the intersection. Our agent after receiving useful parameters such as relative distance and also relative velocities makes the decision whether to apply brake or throttle.  A similar scenario can be applied to pedestrians suddenly  crossing the road. Figure \ref{fig:figure2} explains the proposed scenario.

\section{Deep Reinforcement Learning}
In this section we present the details of our DRL-based braking and throttle system. The aim of reinforcement learning algorithms is to learn a policy which maximizes the expected cumulative award which will be received in the future. The agent performs an action $a_t$ given the state $s_t$ under policy $\pi$. After performing the action, the agent receives a new state and a reward $r_t$. Here the state consists of relative position of the vehicle and the object, i.e $(objpos_x-vehpos_x, objpos_y-vehpos_y)$ and relative velocity of the vehicle and object for the past 10 steps. The action that the agent takes consists of brake or throttle. Brake and Throttle control are both real valued with values in range $[0,1]$
\newline
The initial position of the agent and the objects is fixed for all the simulation.
The values of $v_{veh}$(velocity of agent) and $v_{obj}$(velocity of other car) are set as random so that our reinforcement learning agent learns a wide variety of cases.

For each scenario, there exists a set of events which will result in the termination of the episode. For scenario 1, the episode will end if there is:

\begin{itemize}
    \item \emph{Collision}:  The agent vehicle has breached the safe distance limit of the object.
    \item \emph{Early stop}: The agent vehicle moving stops too early, i.e the static object is very far away from the agent.
\end{itemize}

The events for scenario 2 are mentioned below:
\begin{itemize}
    \item \emph{Collision}:  The agent vehicle has breached the safe distance limit of the object.
    \item \emph{Early Stopping}: Similar to scenario 1, if our vehicle stops very early compared to the distance to the object in the environment.
    \item \emph{High speed at intersection}:  The velocity of our autonomous agent is very high during crossing at intersection. 
\end{itemize}

\subsection{Deep Deterministic Policy Gradient}
Policy methods \cite{sutton2000policy} are preferred for numerous reasons. First, they do not compute the immediate maximum reward and instead rely on total rewards. Secondly, for a continuous action space, such as braking or throttle control, we cannot calculate the Q value \cite{watkins1992q}, of each action, as the action space will grow exponentially, which makes Policy methods more appealing.

Deep Deterministic Policy Gradient is a deterministic, model-free, off-policy method which is used in situations where the action space is continuous. It is inspired by DQN and Deterministic Policy Gradients(DPG).  DQN \cite{mnih2013playing} calculates the Q function for all actions, using a neural network and selects the action with maximum Q value. DPG uses actor-critic style and learns a deterministic policy, i.e. for each state there is a fixed action defined. For continuous action space, we need a differentiable action-value function approximator to avoid calculation of Q value for each action. DDPG is thus an algorithm which combines both the deterministic policy approach and the neural network as function approximators. DDPG also has a target actor and target critic along with actor and critic networks. The important steps in the algorithm are shown as follows. First, the current state is given as input to actor network and $a = \mu (s;\theta^\mu) + N_t$ is computed, where $a$ is the output action for state $s$ and $\theta^\mu$ is the weights of the actor network. $N$ is the random process chosen for exploration of action space. We will be using Ornstein-Uhlenbeck process to add noise to the action output. \cite{PhysRev.36.823}. Next, the updated Q value, i.e $y_i$ is computed as the sum of the immediate reward $r_i$ and the outputs of the target actor and target critic networks for the next state.

\begin{equation}
    \label{e1}
    y_i = r_i + \gamma Q^{'}(s_{i+1}, \mu^{'}(s_{i+1}|\theta^{\mu^{'}})|\theta^{Q^{'}})
\end{equation}

Here, $\gamma$ is the discounting factor. The critic network is updated by minimizing the mean-squared loss between $y_i$ and the original Q value calculated using critic network, i.e $Q(s_i, a_i|\theta^{Q})$. Since DDPG is an off-policy method, for a mini-batch, we take the mean of the sum of policy gradients while updating actor network \cite{lillicrap2015continuous}. The target networks are time-delayed copies of the original networks and they help bring stability to the system by only moving a factor $\tau$ from the original networks at each step, i.e $\theta^{'} \leftarrow \tau \theta + (1-\tau)\theta^{'}, \tau << 1$, where $\theta$ is the original weight of the actor/critic network and $\theta^{'}$ is the weight of the corresponding target network. 

DDPG makes use of a finite buffer to store each step of every episode as a tuple $(s_t, a_t, r_t, s_{t+1})$. These tuples are randomly sampled, thus they remain undifferentiated between episodes during learning \cite{lillicrap2015continuous}.

\subsection{Reward function for Scenario 1}
The reward function is the most important component of the reinforcement learning algorithm. It must be formulated such that it closely resembles the proposed braking and throttle system. The main objective which our control system should achieve are:

\begin{itemize}
    \item Collision must be avoided at all costs
    \item The car should apply control smoothly instead of sudden brake/throttle, so that the driving is more realistic to human driving.
    \item The car should not stop very early as it makes no sense if the obstacle is very far away from the agent.
    \item The car should not cross the intersection at very high speed.
\end{itemize}

To achieve the above objectives, we propose the following reward function for scenario 1:

\begin{equation} \label{eq2}
 r_t=
   \begin{cases}
   -(\alpha distance^2 + \beta)abs(action) - (\eta v_{veh}^2 + \lambda), & \text{if Collision} \\
   -(\alpha distance^2 + \gamma), & \text{if Early Stopping} \\
   +\delta, & \text{otherwise}
   \end{cases}
\end{equation}

where $ distance^2 = (objpos_x-vehpos_x)^2+ (objpos_y-vehpos_y)^2$,  $\alpha, \beta, \eta, \lambda, \gamma, \delta > 0$, $v_{veh}$ is the velocity of the vehicle, and action is the value of throttle or brake taken by agent at any instant of time. 

The first condition is proposed by Chae \cite{8317839}. It suggests that the penalty in case of collision is proportional to the square of the velocity of the vehicle. The second and third part of the reward function are proposed by us. We want the agent to decide when to apply brakes once the obstacle is detected, we don't want our vehicle to stop if the obstacle is far away from the vehicle. So the second condition gives the penalty proportional to the square of the distance between obstacle and vehicle. Finally if we are in a safe state, i.e there is neither collision or early stopping, we simply give a constant positive reward for being in a safe state. In an episode, we want our agent to gather as much positive reward as possible.

\subsection{Reward function for Scenario 2}
The reward function for the scenario where both the agent and the car is moving towards an intersection is:

\begin{equation} \label{eq1}
 r_t=
   \begin{cases}
    -(\alpha distance^2 + \beta)abs(action) - (\eta (v_{veh}-v_{obj})^2 + \lambda), & \text{if Collision} \\
    -(\alpha distance^2 + \gamma), & \text{if Early Stopping} \\
    -(\alpha v_{veh}^2 + \mu),& \text{if High speed at intersection} \\
    +\delta, & \text{otherwise}
\end{cases}
\end{equation}

Here $distance $ is the same as the one defined in scenario 1 and $\alpha, \beta, \eta, \lambda, \gamma, \delta, \mu > 0$. 

The reward function looks very similar to the one proposed for scenario 1. There is one additional condition. In a real-world scenario, we do not drive at very high speed while driving through an intersection. We give a penalty proportional to the square of the velocity of agent. These conditions will ensure that our agent will avoid a collision and also take decisions which closely resemble human driving. The values of weight parameters $\alpha, \beta, \eta$ etc, should be carefully chosen as they serve as a trade-off between each objective. We can assign each of these objectives a certain priority. We give collision avoidance the highest priority, hence, we assign maximum penalty in case of collision. Next comes early stopping and high speed at intersection which are given equal priority. However, the penalty is lower as compared to collision avoidance. Considering this argument, we conclude $ \lambda > \gamma, \mu $.

\section{Experiment}
\subsection{Simulation Setup}
We use CARLA (Car Learning to act), \cite{Dosovitskiy17} an open-source driving simulator which provides an environment to create scenarios commonly encountered by vehicles and allow the agent to learn from the interactions. We simulated our environment by spawning vehicles according to the scenario mentioned in figure \ref{fig:figure1} and \ref{fig:figure2}. In the simulations it is assumed that the relative distance of the other objects like cars, stop sign to the agent is known. For each simulation the agent will be spawned at the same location. The 2 scenarios which we simulated to make models are:

\begin{itemize}
    \item Scenario 1: Static obstacle in front of agent. (Car, Stop sign).
    \item Scenario 2: Dynamic Obstacle, i.e both cars moving towards an intersection.
\end{itemize}

For scenario 2, we also want our agent to start again after emergency braking, so we do not end the episode once the agent avoids the collision. We keep a maximum limit of 7.5 seconds for an episode to end. Time to collision(TTC) is chosen within this maximum limit.

For both the scenarios, the time to collision (TTC) is chosen to be between 1.5 \emph{s} to 5 \emph{s}. Also the safety distance to the obstacle is kept to be 5 \si{\meter}. For scenario 2, a safety distance box is made as shown in figure \ref{fig:figure2}.  

The velocity of agent is sampled from uniform distribution, with $v_{init}^{min} = 8.33 \si[per-mode=symbol]{\meter\per\second}  (30 \si[per-mode=symbol]{\km\per\hour})$ and $v_{init}^{max} = 27.77 \si[per-mode=symbol]{\meter\per\second} (100 \si[per-mode=symbol]{\km\per\hour})$. The wide range of values will make our agent learn to apply brake/throttle in most extreme cases also. 
The detailed simulation setup for both of these scenarios is shown in Table \ref{tab:simulation}


\begin{table}
\caption{Simulation Setup for experiment}
\label{tab:simulation}
\subfloat[Scenario 1]{
\begin{tabular}{ |c|c|}
\hline
$v_{init}$ of agent & U(8.33, 27.77) \si[per-mode=symbol]{\meter\per\second} \\
\hline

$v_{init}$ of obstacle & 0 \si[per-mode=symbol]{\meter\per\second} \\
\hline

Rel. Position of obstacle & 60 \si[per-mode=symbol]{\meter} \\
\hline
Safety Distance & 5 \si[per-mode=symbol]{\meter} \\
\hline
Throttle & $\in$ ${\rm I\!R}$ with values [0,1] \\
\hline
Brake & $\in$ ${\rm I\!R}$ with values [0,1] \\
\hline
$\Delta T$ & 0.1 \emph{s} \\
\hline
\end{tabular}}
\quad
\subfloat[Scenario 2]{
\begin{tabular}{ |c|c|}
\hline
$v_{init}$ of agent & U(8.33, 27.77) \si[per-mode=symbol]{\meter\per\second} \\
\hline

$v_{init}$ of obstacle & U(8.33, 27.77) \si[per-mode=symbol]{\meter\per\second} \\
\hline

Position of agent from intersection & 45 \si[per-mode=symbol]{\meter} \\
\hline

Position of obstacle from intersection & 45 \si[per-mode=symbol]{\meter} \\
\hline

Safety Distance & 5 \si[per-mode=symbol]{\meter} \\
\hline
Throttle & $\in$ ${\rm I\!R}$ with values [0,1] \\
\hline
Brake & $\in$ ${\rm I\!R}$ with values [0,1] \\
\hline
$\Delta T$ & 0.1 \emph{s} \\
\hline
\end{tabular}}
\end{table}

\subsection{Training}
The neural network architecture consists of a fully connected feed-forward neural network with 5 hidden layers for both the actor and critic networks. We use the same architecture for both of the scenarios, the difference is in their reward functions. Adam optimizer is used for both actor and critic networks. \cite{kingma2014adam}. A total of 2000 episodes have been trained for both scenarios.The details of the architecture is shown below:

\begin{itemize}
    \item State buffer size : $\emph{n} = 10$
    \item Network architecture - Fully connected feed-forward network with 5 hidden layers
    \item Number of nodes for each layers: [40(Input layer), 400, 200, 100, 200, 400, 1(Output Layer)]
    \item Non linear Function - leaky ReLU.
    \item Actor learning rate - 0.00005
    \item Critic learning rate - 0.0005
    \item Replay buffer size - 20000
    \item Minibatch size - 16
    \item Gamma - 0.99
    \item Tau - 0.001
    \item Reward function - $\alpha=0.01$, $\beta=0.1$, $\eta=0.01$, $\lambda=50$, $\gamma$ (Scenario 1) = 15, $\gamma$ (Scenario 2) = 20, $\mu=30$ ,$\delta=0.5$.
\end{itemize}

There is one neuron at the output. We have used tanh activation layer at the output, so that the network outputs the value between -1 and 1. If the output is less than 0, we apply the brake and if the value is greater than 0, we apply throttle.

\subsection{Results}
\begin{figure}[h!]
    \centering
    \includegraphics[scale=0.7]{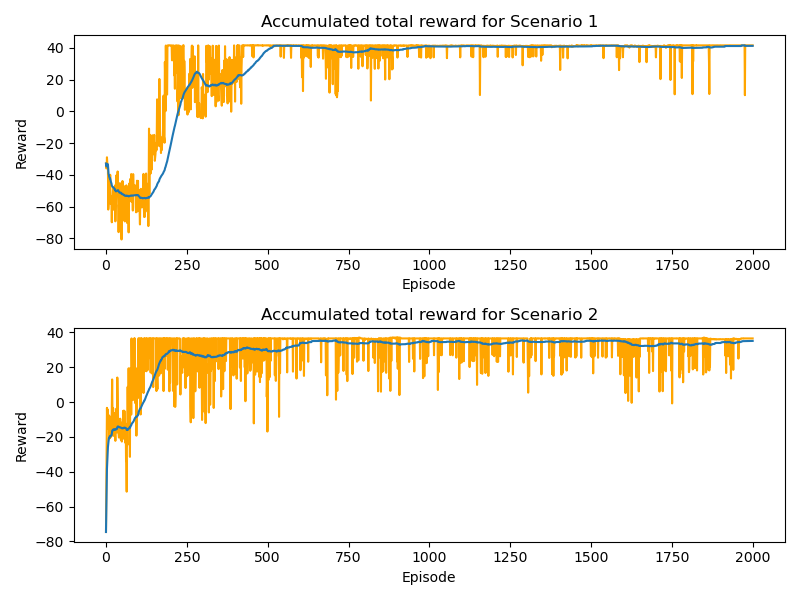}
    \caption{Value function during training}
    \label{fig:figure3}
\end{figure}

Figure \ref{fig:figure3} depicts the accumulated reward for scenario 1 and 2. For scenario 1,we can see that our model makes quite an improvement in the accumulated reward after 200 episodes, from 200 episodes to 700-750 episodes, it follows an increasing curve and later the curve remains more or less the same. Occasionally the curve shows some small spikes which is expected since we are sampling the initial velocity of a car to be 100 km/hr which is very high and even if we apply full braking action from the start of the episode, we cannot avoid the collision. 

\begin{figure}[h!]
    \centering
    \includegraphics[scale=0.7]{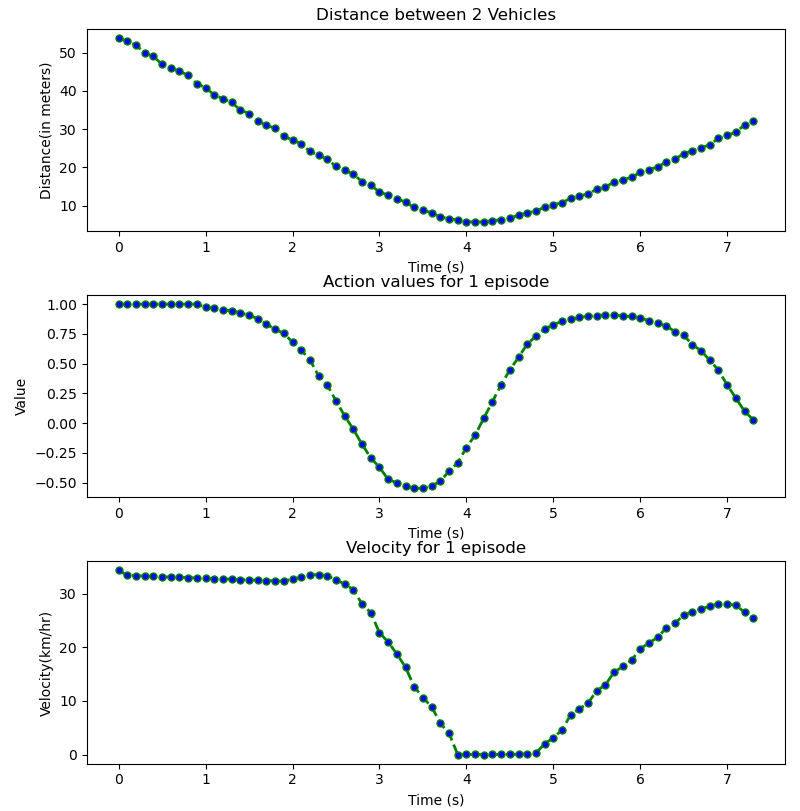}
    \caption{Trajectory of action and velocity of agent during 1 episode}
    \label{fig:my_label}
\end{figure}

The accumulated reward curve for scenario 2 follows a similar trend as scenario 1. But there are more spikes as compared to scenario 1. This can be attributed to scenario 2 being more complex and the agent needing to take into account the high speed at intersection, which makes it hard to learn a good policy. Now we consider one test episode of scenario 2. Figure \ref{fig:my_label} shows the velocity of the car as the agent takes braking or throttle decisions during one episode. We observe that as the car approaches the intersection, the agent slowly decreases throttle, so there is very less decrease in velocity, but after 3 seconds, it detects the object crossing the intersection at high speed, so it applies brakes till the vehicle reaches a speed of 0. Once the car avoids an emergency situation (collision), it again applies throttle to speed up the car. This explains the valley like curve of velocity in figure \ref{fig:my_label}, between 2 and 6 seconds of the episode. At last, when the vehicle achieves desired velocity, it reduces the throttle input to make it constant. This curve closely resembles human driving while crossing intersections. The inclusion of early stopping and avoiding high speed at intersections ensures the car is always in a safe state. Since we are using real values of brake and throttle as compared to previous study by Chae \cite{8317839} where the author proposes discrete values of deceleration, we get a smooth decrease/increase in velocity. Also the velocity is in the safety limits as we have imposed penalties for high speed.

\section{Conclusion and Future Work}
We have demonstrated the braking and throttle control system for 2 scenarios. One where braking is very important, while the other scenario needs both braking and throttle action. DDPG is used to ensure that the values of brake and throttle are smoothly changing instead of a sudden change. We have proposed measures to avoid early stopping and high speed movement at an intersection, making autonomous driving similar to human driving. This work can be further extended to other such situations, where the vehicle must apply brake or throttle to avoid emergency situations. An interesting study would be to see the effect of braking and throttle actions along with steering of the vehicle in these situations. One can create a unified model for all these scenarios with a singular reward function. Also, the effect of change of weather conditions can also be considered so as to make the model more realistic.

\bibliographystyle{unsrt}  
\bibliography{references}  
\end{document}